\documentclass{article}

\usepackage[final]{midl_2018}

\usepackage[utf8]{inputenc}          
\usepackage[T1]{fontenc}             
\usepackage{hyperref}                
\usepackage{url}                     
\usepackage{graphicx,booktabs,array} 
\usepackage{amsfonts}                
\usepackage{nicefrac}                
\usepackage{microtype}               
\usepackage{natbib}
\usepackage{hyperref}
\usepackage[printwatermark]{xwatermark}
\usepackage[tbtags]{amsmath}
\usepackage[mathscr]{euscript}
\usepackage[capitalise]{cleveref}
\usepackage{multirow}
\usepackage{lineno}
\usepackage{amsfonts,dsfont,bm}

\providecommand{\ve}[1]{{\bm{#1}}} %
\providecommand{\mat}[1]{{\bm{#1}}} %

\DeclareMathOperator{\igual}{ \negthinspace=\negthinspace}

\newlength\figureheight
\newlength\figurewidth
\newcommand{\addpic}{11em}
\newcommand{\addpich}{9em}
\title{Simultaneous synthesis of FLAIR and segmentation of white matter hypointensities from T1 MRIs}
\author{
  Mauricio Orbes-Arteaga \\
  Translational Imaging Group\\
  University College London\\
  Biomediq A/S\\
  Copenhagen Ø, Denmark\\
  \texttt{henry.arteaga.17@ucl.ac.uk} \\
  \And
   M. Jorge Cardoso\\
  Translational Imaging Group\\
  University College London\\
  \texttt{m.jorge.cardoso@ucl.ac.uk} \\
  \And
  Lauge Sørensen\\
  Biomediq A/S  \\
  Department of Computer Science\\
  University of Copenhagen, \\
  \texttt{lauges@di.ku.dk} \\
  \And 
  Marc Modat\\
  Translational Imaging Group\\
  University College London\\
  \texttt{m.modat@ucl.ac.uk} \\ 
  \And
  Sébastien Ourselin\\
  Translational Imaging Group\\
  University College London\\
  \texttt{s.ourselin@ucl.ac.uk} \\
  \And
  Mads Nielsen\\
  Biomediq A/S  \\
  Department of Computer Science\\
  University of Copenhagen, \\
  \texttt{	madsn@di.ku.dk} \\
  \And 
  Akshay Pai\\
  Biomediq A/S  \\
  Department of Computer Science\\
  University of Copenhagen, \\
  \texttt{akshay@di.ku.dk} \\
}

\begin{document}
\maketitle
\begin{abstract}
Segmenting vascular pathologies such as white matter lesions in Brain magnetic resonance images (MRIs) require acquisition of multiple sequences such as T1-weighted (T1-w) --on which lesions appear hypointense-- and  fluid attenuated inversion recovery (FLAIR) sequence --where lesions appear hyperintense--. However, most of the existing retrospective datasets do not consist of FLAIR sequences. Existing missing modality imputation methods separate the process of imputation, and the process of segmentation. In this paper, we propose a method to link both modality imputation and segmentation using convolutional neural networks. We show that by jointly optimizing the imputation network and the segmentation network, the method not only produces more realistic synthetic FLAIR images from T1-w images, but also improves the segmentation of WMH from T1-w images only. 

\end{abstract}

\section{Introduction}
Segmenting white matter hyperintensties/hypointensities (WMH) from brain magnetic resonance images (MRI) have a profound impact in understand the role of vascular pathology in various neurological disorders \cite{caligiuri2015automatic}. Segmenting WMH manually is not feasible due to time, and inter/intra-rater variability. 

In the recent years, deep learning strategies have gained attention in medical image analysis. Specially, convolutional Neural Networks(CNN) have been widely used in disease classification, segmentation, and registration tasks \cite{litjens2017survey,kamnitsas2017efficient,havaei2016hemis}. Particularly, CNNs have also become the first choice in the segmentation of WMH. For example, the top-three performing methods in the WMH segmentation challenge~\footnote{\url{http://wmh.isi.uu.nl}}, have relied on some form of CNNs.

From a more practical perspective, segmenting vascular pathologies such as WMH usually requires multiple MRI modalities \cite{griffanti2016bianca,dadar2017validation}. Most often, in addition to the usual T1 scans, FLAIR sequences are also obtained since additional MRI sequences are specifically designed to provide complementary information to T1 scans. Having mentioned this, it is notable that most of the existing datasets only contain T1 scans or T1/T2/PD scans due to logistical reasons. Given the presence of limited data with desired multiple modalities, data imputation methods are used to learn the synthesis of missing modality using T1 scans. The intention of imputing data is to guide the optimization using prior information, i.e., the available FLAIR sequence. As stated in \cite{van2015does}, synthetic data helps the segmentation because of two reasons. Firstly, the flexibility of synthesis model allows finding features that can not be seen by the classifier in an otherwise single-modality model. Secondly, the size of the training set is synthetically increased which is useful in the training process.

Among CNN-based imputation methods, the most popular ones using a flavor of generative adversarial networks (GANs) \cite{goodfellow2014generative}. For instance,  ~\cite{nie2017medical} use GANs to generate CT images from MRI images. However, most of the current implementations treat synthesis as a preprocessing step \cite{ben2018cross,zhang2018translating,huo2017adversarial}. This restricts the network, and the features may not be particularly useful for the final segmentation.

In this paper, we proposed a simultaneous training based synthesis method that combines generation of the missing modality and segmentation -- inspired from ~\cite{tran2017bayesian}. Experiments on the WMH segmentation challenge 2017 dataset shows that using the proposed method to synthesize FLAIR images, we not only obtain higher quality synthetic flair images (when compared to treating synthesis a preprocessing step) but also improve the segmentation of WMH using T1-w images only.

\section{Methods}
Let $\mathscr{X}\igual\{ \ve{\mat{X}^n,L^n}:1,\dots,N\}$ be an annotated training set which have $N$ subjects . Here, $\mat{X}=\{X_a,X_b\}$, is a pair of MRI images from two different modality sources for a given subject, and $\mat{L}$ is a volume with the manual annotation for WMH. The goal in multi modal segmentation task is to find  a mapping $C(\mat{X},\theta_c)$ from a pair of available modalities to a corresponding segmentation. 

\begin{equation} \label{eq:eq_1}
C: \{X_a,X_b\}, \rightarrow  L
\end{equation}

Here, $C$ is a function represented by a CNN with parameters $\theta_c$. We then train $C$ to maximize: 

\begin{equation} \label{eq:eq_2}
\max_{\theta_c} \mathbb{E} [ \log \ p(L|X_a,X_b,\theta_c) ] 
\end{equation}

\noindent
It is evident that to train, and subsequently test such a scheme, both modalities are needed. This is a restriction, specially when the network is used to test retrospective data with missing modalities. One of most common approaches to deal with missing modalities is to impute them. Formally, a function $G$ (a CNN) is trained to learn a mapping between the available modality and the missing modality., i.e $G(X_a) \approx X_b$. Subsequently, the synthesized modality is used in conjunction with the available modality to train a classifier for segmentation. The optimization function for the classifier in~\Cref{eq:eq_2} can be re-written as:
\begin{equation} \label{eq:eq_4}
\max_{\theta_c} \mathbb{E} [ \log \ p(L|X_a,G(X_a),\theta_c)  ] 
\end{equation}

Note that in this scheme, the generation and the classification are different optimizations. No complementary information is taken into account. Therefore, in this work, we aim to learn the generation and classification (respectively performed by $G$ and  $C$ ) simultaneously so that $C$ reinforces the generation $G$ to produce not only realistic images but also relevant features that help in the optimization of $C$.

\begin{figure}[ht]
  \centering
      \includegraphics[width=0.7\textwidth]{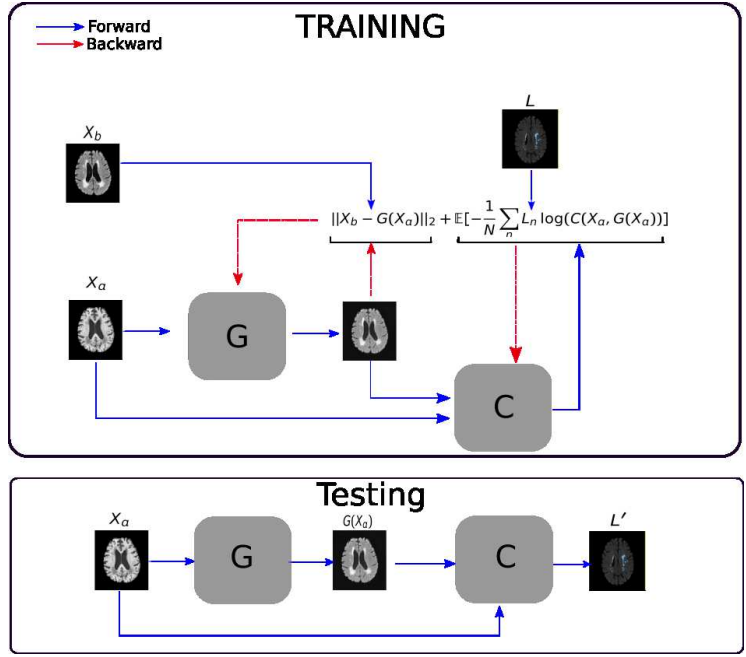}
  \caption{Illustration of process follow for training and testing of our method. }
  \label{fig:architecture}
\end{figure}

The scheme is basically composed of two networks, a generator $G$ and a classifier $C$ where both networks are trained end to end iteratively, see ~\Cref{fig:architecture}. The classifier training is linked to the generator by taking both  the real T1 image denoted by $X_a$ and the generated image $G(X_b)$ to produce a segmentation $\mat{L'}$.  The loss function of the classifier network is:

\begin{equation}\label{eq:5}
\mathscr{L}_C =  \mathbb{E} [- \frac{1}{N} \sum_n L_n \log (C(X_{a}^{n},G(X_{a}^{n}))] 
\end{equation}

In order to train $G$ to produce images looks like as FLAIR, we use L2 as a reconstruction error between the real missing modality image and its corresponding generation. One may then view the classifier to be a regularization term to the generator or vice versa. The L2 loss for the generator is given by:

\begin{equation}\label{eq:6}
\mathscr{L}_G =  ||X_b-G(X_a)||^2 +  \mathbb{E} [- \frac{1}{N} \sum_n L_n \log (C(X_{a}^{n},G(X_{a}^{n}))] 
\end{equation}

\subsection{Network architectures}
We use U-Nets~\cite{li2018fully} (winner in 2017 MICCAI- WMHs segmentation challenge) as the segmentation network, and a modification of it as a generation network. The changes involve changing the number of inputs channels from two to one which corresponds to the T1 modality, we also change the \verb|Sigmoid| function in the final layer by \verb|LeakyRelu|. We use Adam optimizer with learning rate \verb|0.0002| for both the networks, and batch normalization. The classifier and generator are trained iteratively with the same frequency. We do not use any data augmentation.

\section{Experiments and results}
\subsection{Data and Experiment}
We validated our proposed method on the training dataset from the 2017 White Matter Hyperintensity Segmentation Challenge (\url{http://wmh.isi.uu.nl}). This dataset is composed of T1 and FLAIR scans for 60 subjects from three different clinics (Utrecht, Singapore, and AmsterdamGE3T, 20 subjects for each one), the data is complemented with manual annotations of WMH from presumed vascular origin. FLAIR images have been used as a reference for label annotations, so, T1 images have been registered to this space.  The images were also corrected for bias field inhomogeneities using SPM12. 
\noindent
As a further preprocessing we use only two of three stages performed in \cite{li2018fully}, which include \textit{i)} cropping or padding of axial slices \textit{ii)} Gaussian normalization of voxel intensities.We did not perform data augmentation as these did not show significant improvement in segmentation.

All the methods were evaluated using a 6-fold cross validation. The dataset was split in such a way that all the 60 images are tested at least once. For each fold, we pick 10 subjects for test, 5 for validation, and the remaining 45 are used for training. For evaluation, dice scores (DSC), false positive rates (FPR), and false negative rates (FNR) are used. 

\subsection{Results} \label{segRes}
\begin{figure}[ht]
  \centering
      \includegraphics[width=1\textwidth]{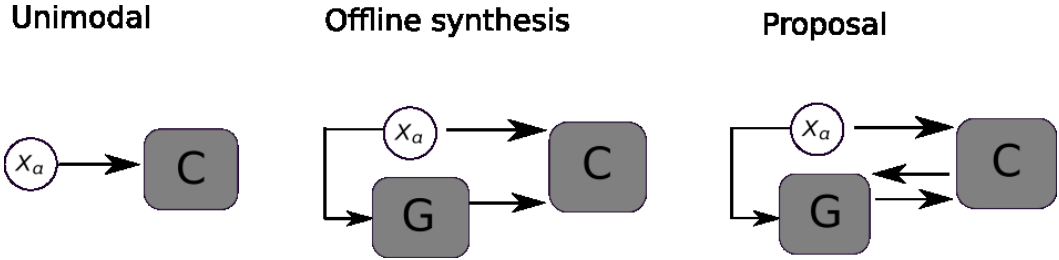}
  \caption{Illustration of methods of comparison, $X_a$ represent a T1 image.}
  \label{fig:Baselines}
\end{figure}

We evaluated our method in segmenting WMH from T1-w images using: a) Synthesized FLAIR images by treating the synthesis as a preprocessing step -- we will refer to this method as \emph{offline synthesis}; b) Synthesized FLAIR images using the proposed method, and c) without any synthesis -- we will refer to this method as \emph{Unimodal}. Baseline methods are illustrated in \Cref{fig:Baselines}

~\Cref{sample-table} shows the mean of each measure for all considered methods. As we can see, our method achieves higher dice scores than baseline methods. A mean dice improvement of nearly three percent is obtained using our proposed method when compared the baseline method without any imputation. In addition, the proposed method also improves segmentation when compared to an offline synthesis. 

 \begin{table}[ht]
 \centering
  \caption{  Average of performance measures for all comparison methods, results in bold are significantly different  (p<0.005) from the baseline \textit{Unimodal} method (top row)}
  \label{sample-table}
  \begin{tabular}{lccc}
    \toprule
    &\multicolumn{3}{c}{Evaluation Metric}             \\
    \cmidrule{2-4}
    Method     &  DSC($\%$)    &  FPR($\%$) & FNR($\%$) \\
    \midrule 
    Unimodal              & 55.99           &   78.67            &  38.06 \\
    Offline synthesis     & 54.42           &   63.50            &  \textbf{43.39} \\     
    Proposal              & \textbf{57.81}  & \textbf{58.20}     &  \textbf{41.33} \\
    \bottomrule
  \end{tabular}
\end{table}

It is important to note, that our proposed method shows a FPR  $20.47 \%$ lower than  \textit{Unimodal} and $5.3\%$ lower than \textit{offline synthesis} method, showing the effectiveness of our method to reduce the number of false positives. On the other hand, \textit{Unimodal} method shows the lower rates in terms of FN. 

\begin{table}[!h]
\begin{tabular}{cccc}
\rotatebox{90}{ \ \ \ \ \ \ \ \ FLAIR}  & \includegraphics[width=\addpic,height=\addpich]{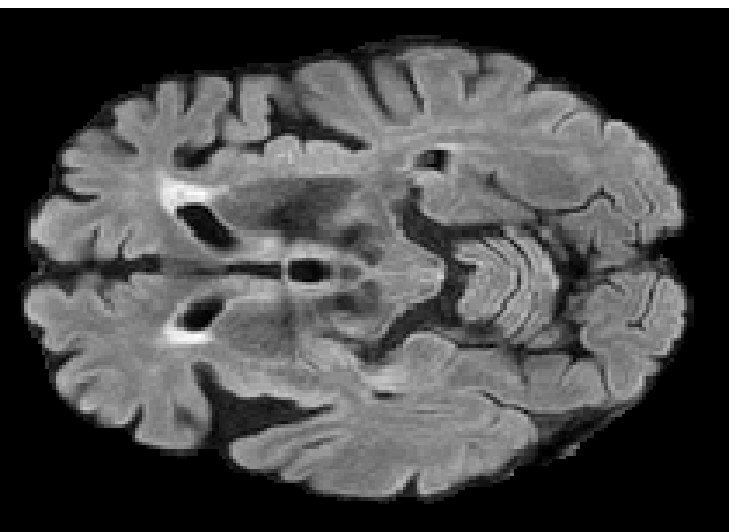}  & \includegraphics[width=\addpic,height=\addpich]{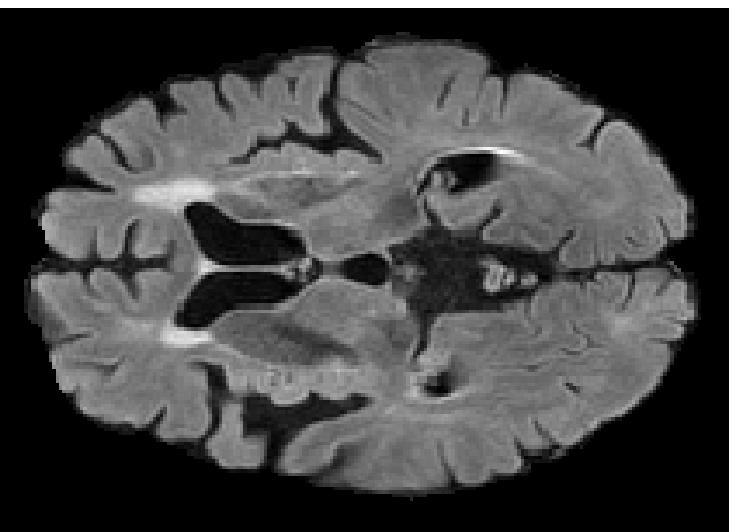}  & \includegraphics[width=\addpic,height=\addpich]{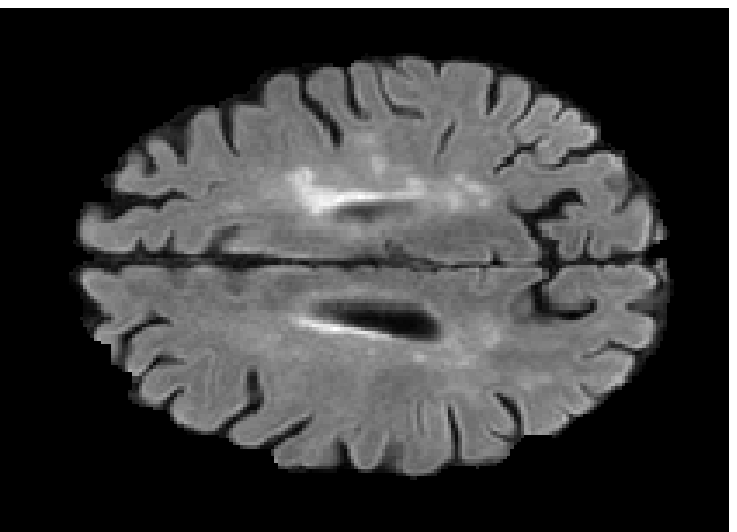} \\
\rotatebox{90}{ \ \ \ \ \ \ \ \ T1}  & \includegraphics[width=\addpic,height=\addpich]{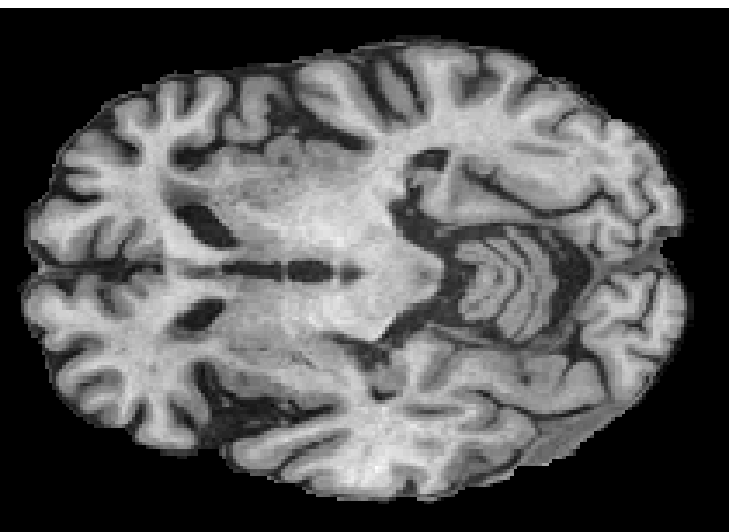}  & \includegraphics[width=\addpic,height=\addpich]{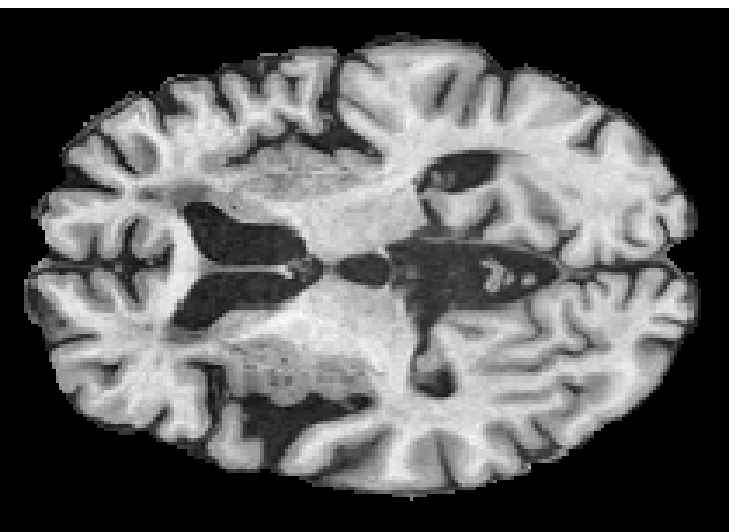}  & \includegraphics[width=\addpic,height=\addpich]{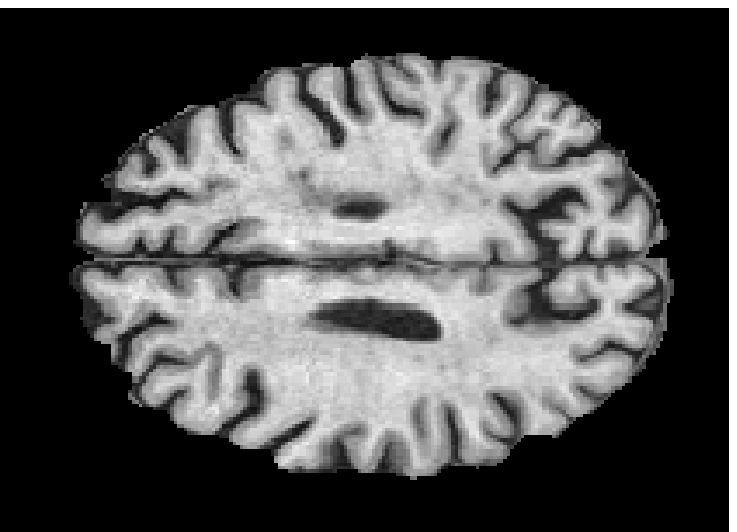} \\
\rotatebox{90}{ \ \ \ \ \ \ \ \ Ground Truth}  & \includegraphics[width=\addpic,height=\addpich]{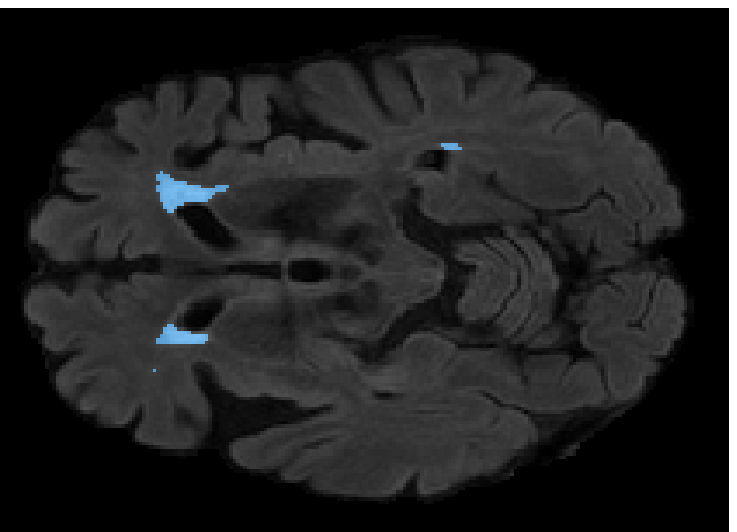}  & \includegraphics[width=\addpic,height=\addpich]{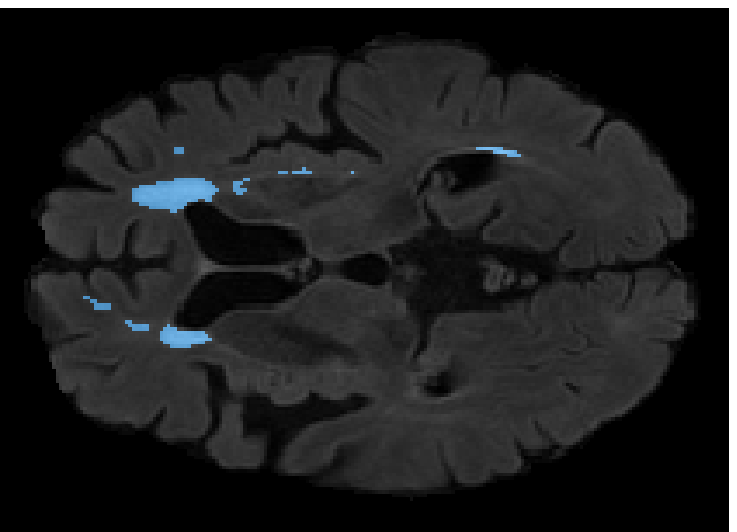}  & \includegraphics[width=\addpic,height=\addpich]{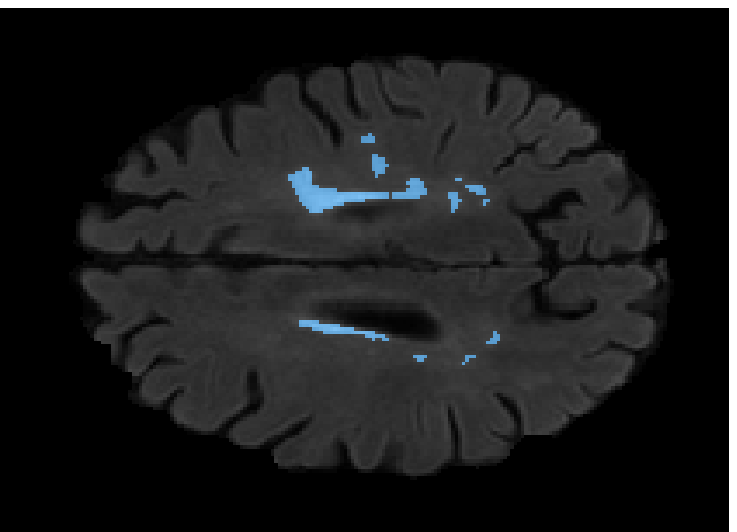} \\
\rotatebox{90}{ \ \ \ \ \ \ \ \ Unimodal}  & \includegraphics[width=\addpic,height=\addpich]{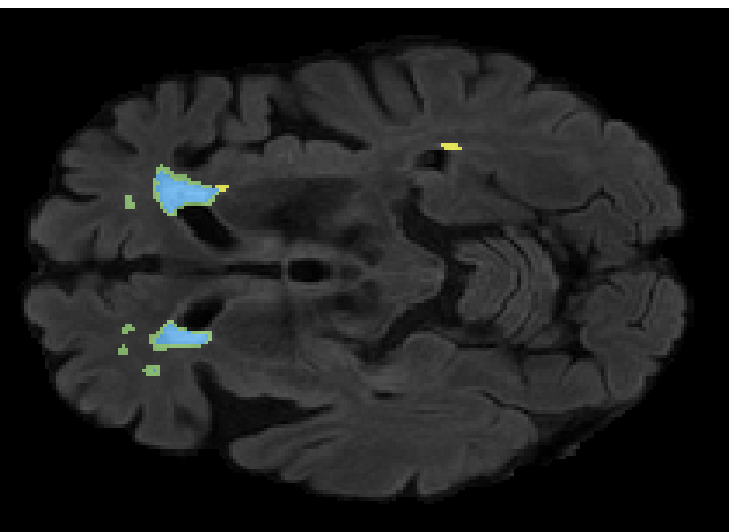}  & \includegraphics[width=\addpic,height=\addpich]{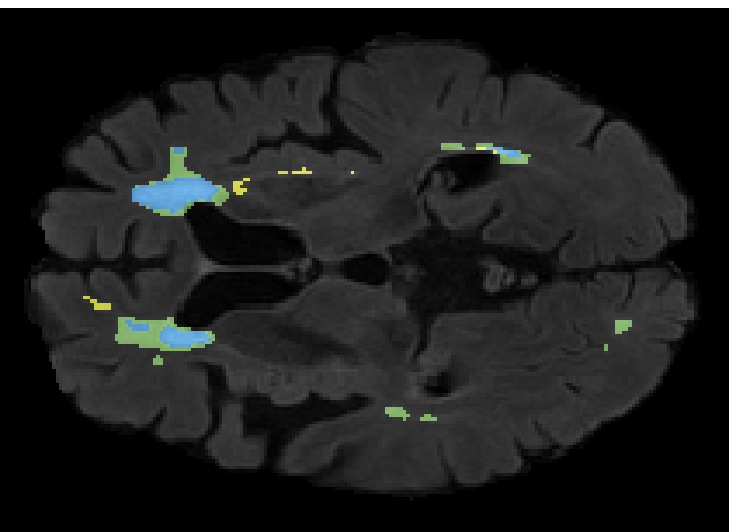}  & \includegraphics[width=\addpic,height=\addpich]{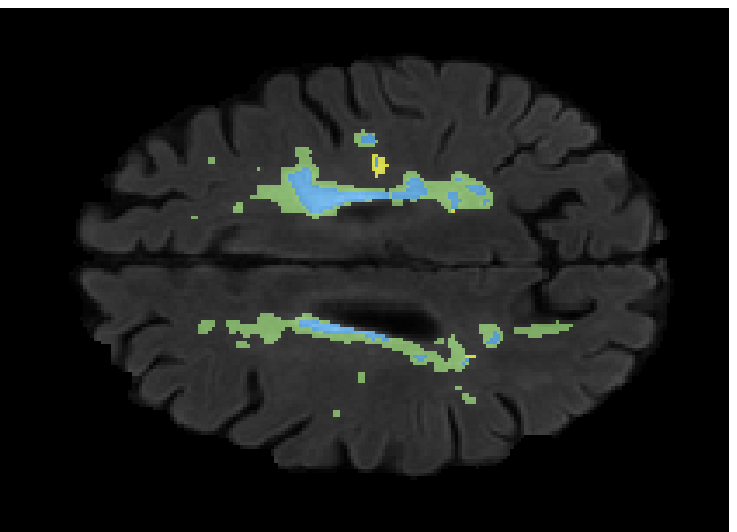} \\
\rotatebox{90}{ \ \ \ \ \ \ \ \  Offline synthesis}  & \includegraphics[width=\addpic,height=\addpich]{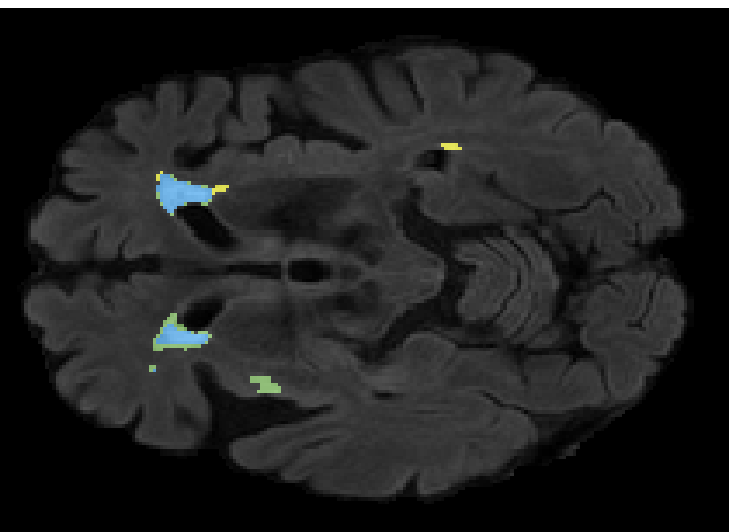}  & \includegraphics[width=\addpic,height=\addpich]{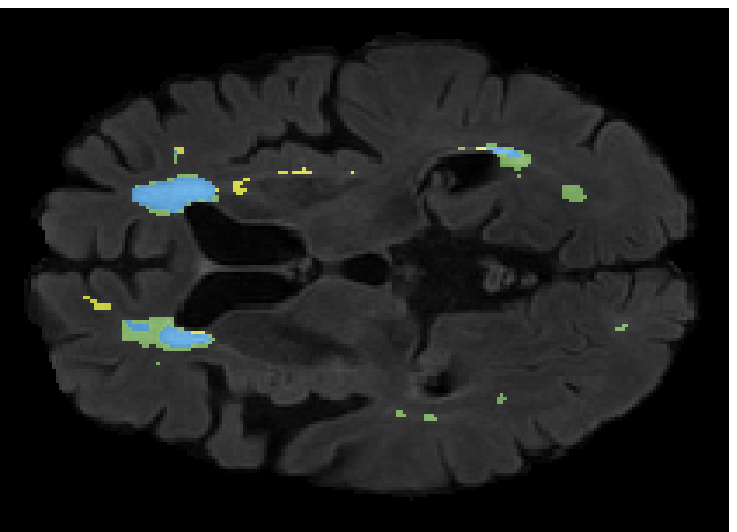}  & \includegraphics[width=\addpic,height=\addpich]{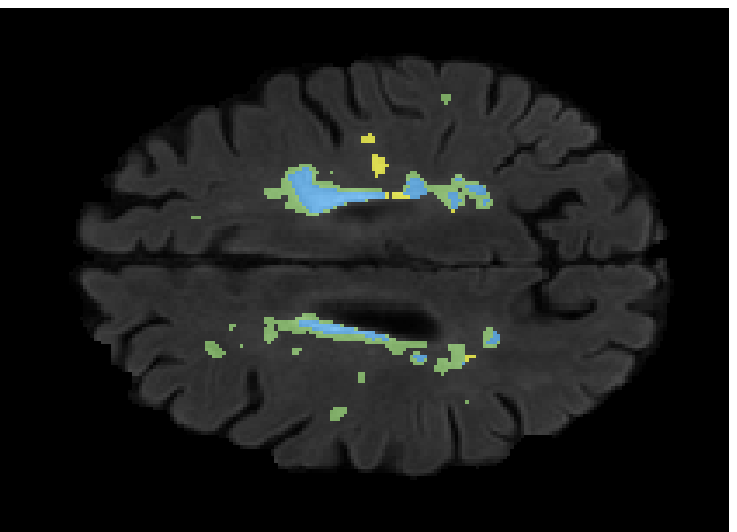} \\
\rotatebox{90}{ \ \ \ \ \ \ \ \ Proposal}  & \includegraphics[width=\addpic,height=\addpich]{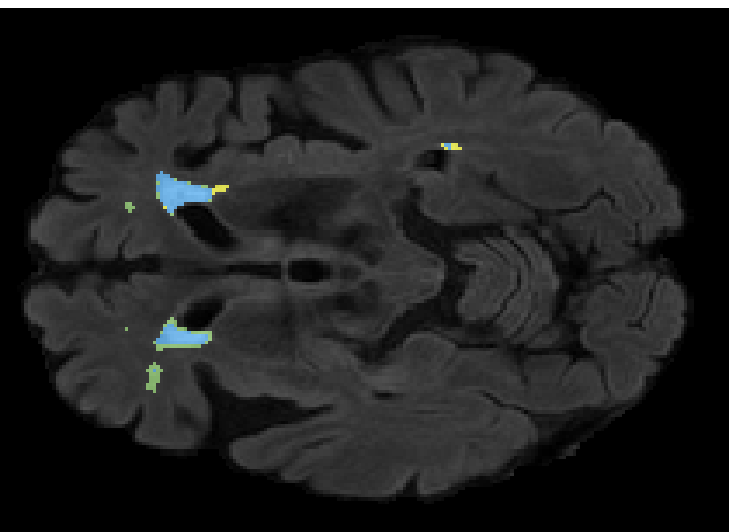}  & \includegraphics[width=\addpic,height=\addpich]{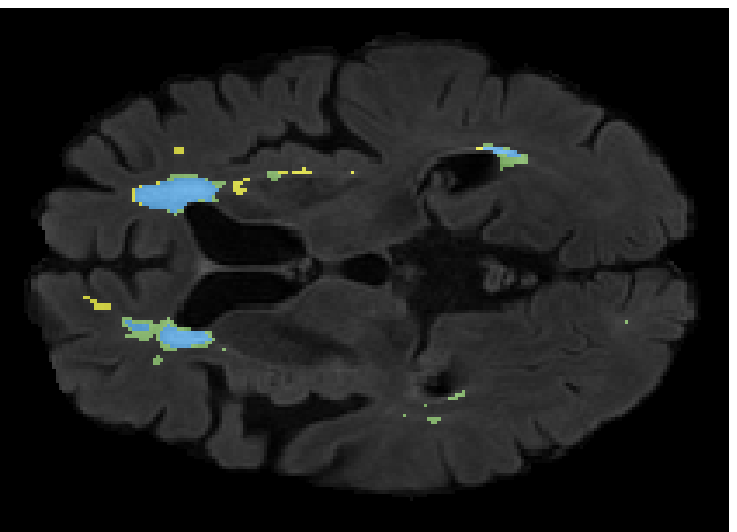}  & \includegraphics[width=\addpic,height=\addpich]{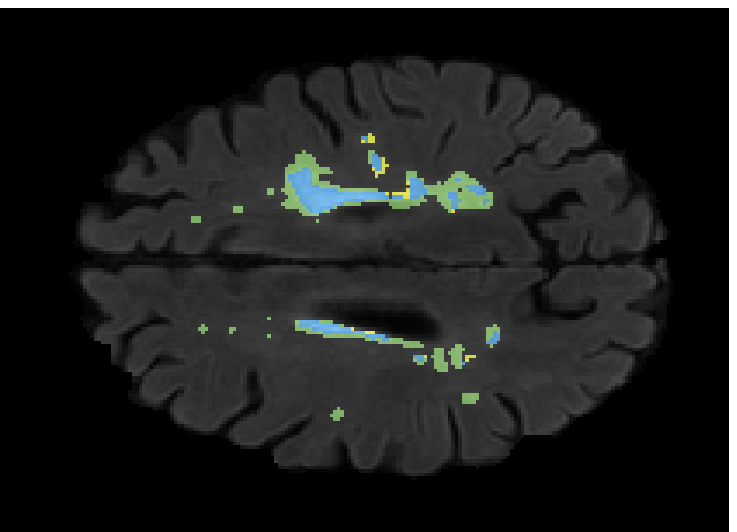} \\
\end{tabular}
\caption{Segmentation results for all proposed methods, each column represent a different slide in the image, blue areas are regions which were correctly labeled, false positives are shown in green, and false negatives in yellow}\label{fig:QualitativeSegmentations}
\end{table} 

In order to better understand the above results, we visually analyzed the output segmentation performed for each method.  \Cref{fig:QualitativeSegmentations} shows the results for three different slices (one slice per column). As illustrated, the proposed method is able to produce less false positives.  It is also important to note that, unimodal segmentation is the one that produces more false positives, showing the advantage of using synthetic data. Regarding the nature of false positives, it can be easy to see in the third column a large number of false positives are on the border of periventricular lesions for the \textit{Unimodal} method in comparison to the proposed method. Also from the first and second column, it can be observed that \textit{Unimodal} tend to produce more small regions of false positives near to cortical areas.  Removing such false positives requires additional post-processing steps, therefore, it is of value avoid this kind of over-segmentation. It can also be noted that synthesis methods tend to produce the same kind of false negatives, this may be due to the blurring effects in synthesized images since the information available during testing is limited -- which otherwise is available from a FLAIR sequence. 

\subsection{Results of Generation}\label{resGen}
Here we compare the generate FLAIR images obtained for the generator using our optimization strategy against the generated images obtained for using \textit{ off-line  synthesis}. Firstly, images were quantitatively evaluated in terms of reconstruction using two well known measures, namely mean absolute error(MAE) and peak-signal-to-noise-ratio(PSNR). Results of reconstruction measures are shown in  \Cref{Generation-results} , as we can see our proposal outperforms the baseline approach in both MAE and PSNR. Specifically, images generated for our proposed method achieve an average PSNR of 11.01 which is considerably higher compared with 9.65 obtained for images generated \textit{Offline}. Reconstruction superiority of our methods is confirmed by the MEA results, 0.26 and 0.31 for our proposal and the baseline respectively. 

\begin{table}[!ht]
  \caption{ Average MAE and PSNR between real Flair images and the synthetic images generated for each method}
  \label{Generation-results}
  \centering
  \begin{tabular}{lcc}
    \toprule
    & \multicolumn{2}{c}{Method}             \\
    \cmidrule{2-3}
    Measure     &  Offline synthesis   &  proposal  \\
    \midrule 
    MAE         &   0.3153      &  0.2566    \\
    PSNR(DB)        &  9.65         &  11.01     \\
    \bottomrule
  \end{tabular}
\end{table}

In order to analyze qualitatively the results of our generator, we extract slices with different WMHs loads, \cref{fig:Reconstructions} shows the reconstruction results for three different levels of loads. As we can see in the first row, both methods produce a similar response in regions with a low load of lesions, it can be observed that generated images are similar to the real FLAIR images in the left, and these not present evident structural distortions. However, it can be noted images exhibit blurred effects, which can be due to L2 based optimization, more complex generative networks with adversarial loss optimization as GANs tend to eliminate blurred effect but at the expense to produce structural distortions. In the application presented in this work it is important to preserve the structural information, thus, our L2 based optimization present a good balance between preserve structural information and blurred effects. In the second and third column, it can be observed the performance of both methods when facing the presence of lesions, as can be seen, both methods have a good response to large and contiguous lesions. It also can be noted  both methods tend to produce poor performance in small and diffuse WMHs marked in red, note, these lesion do not exhibit identifiable patterns in T1 images, however it can be seen that our proposed method is more sensitive to these patterns which enable to highlight some small regions as those marked in green. 
 

\begin{table}[!ht]
\begin{tabular}{cccc}
FLAIR & T1 & Offline synthesis & Proposal \\
\includegraphics[width=9em]{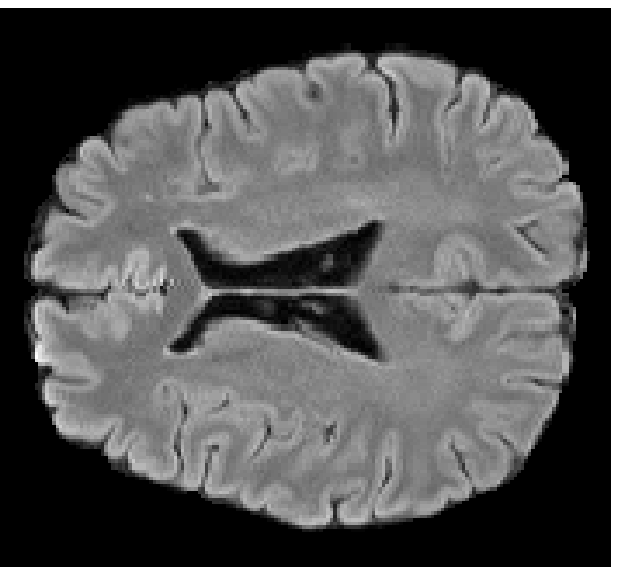}  & \includegraphics[width=9em]{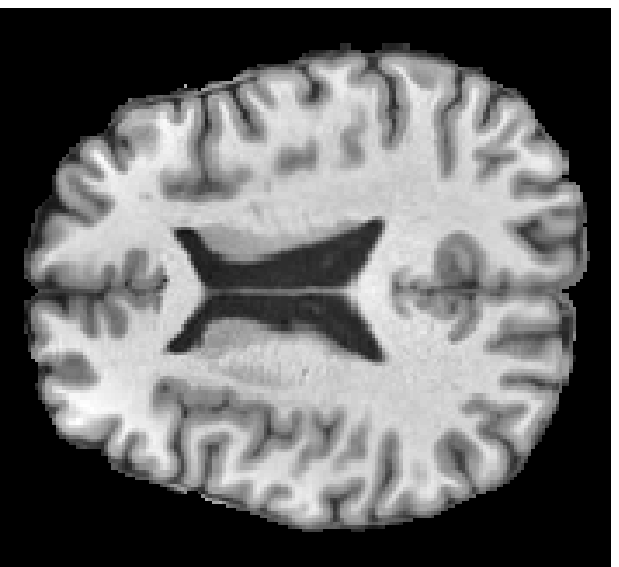}  & \includegraphics[width=9em]{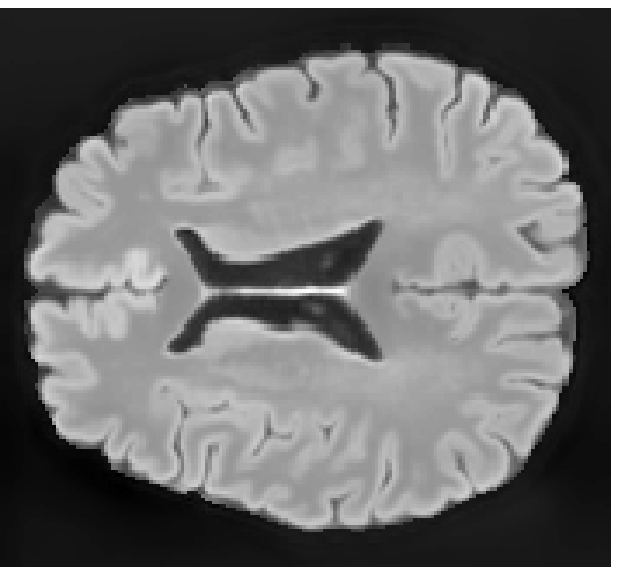} &  \includegraphics[width=9em]{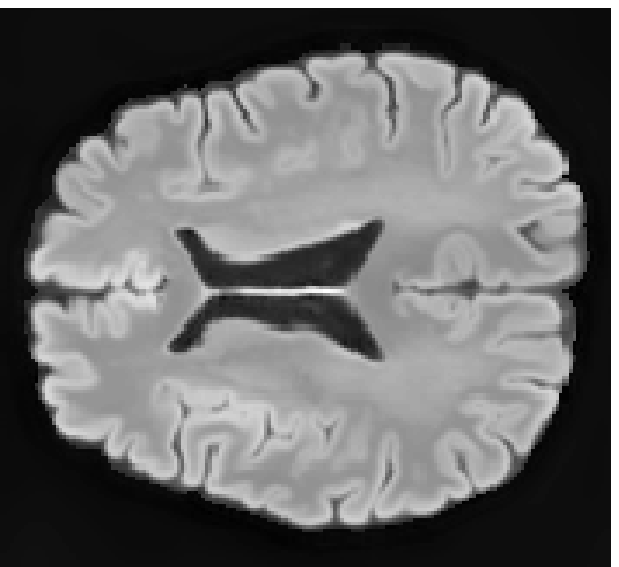}\\
\includegraphics[width=9em]{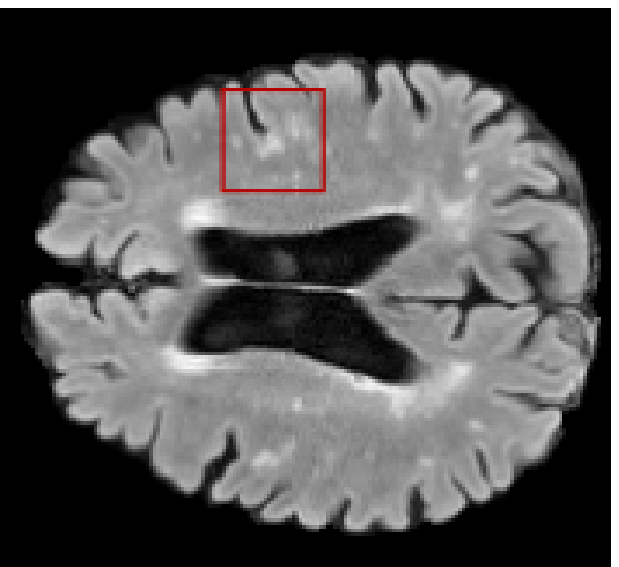}  & \includegraphics[width=9em]{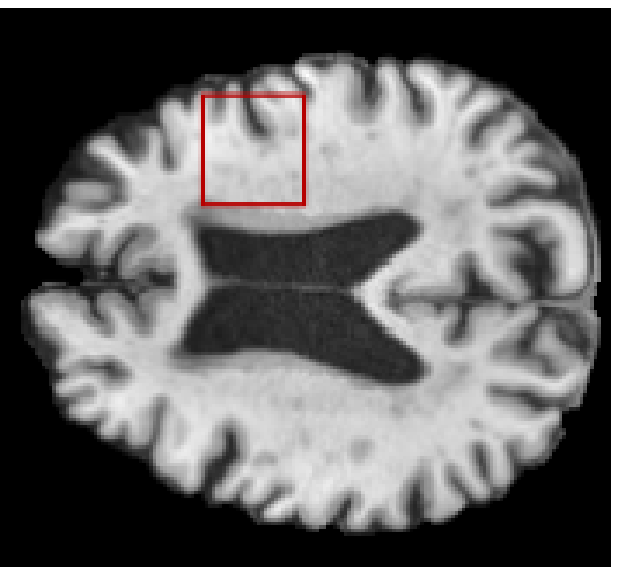}  & \includegraphics[width=9em]{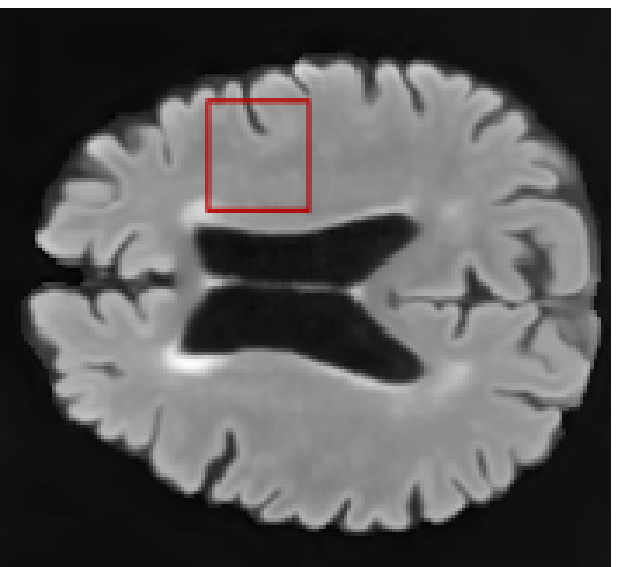} &  \includegraphics[width=9em]{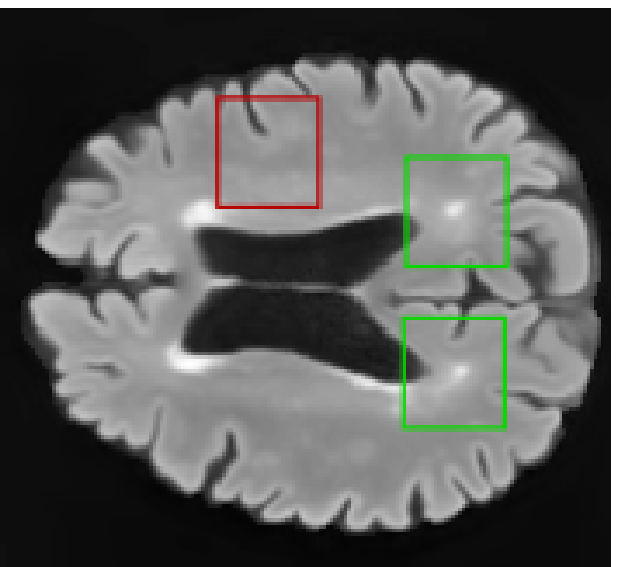}\\
\includegraphics[width=9em]{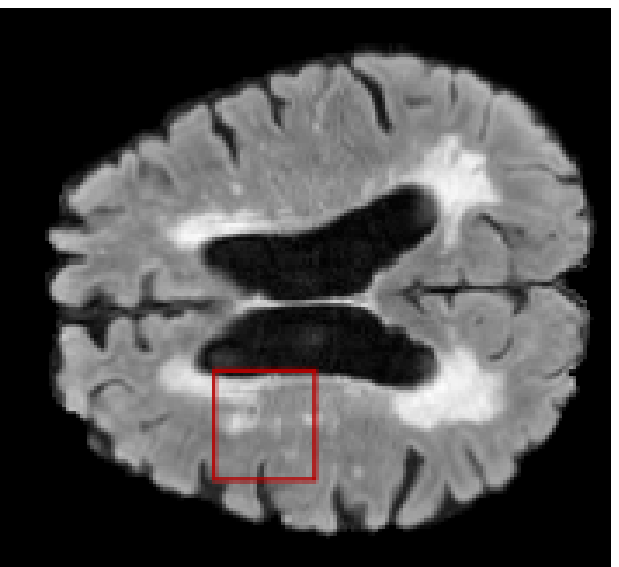}  & \includegraphics[width=9em]{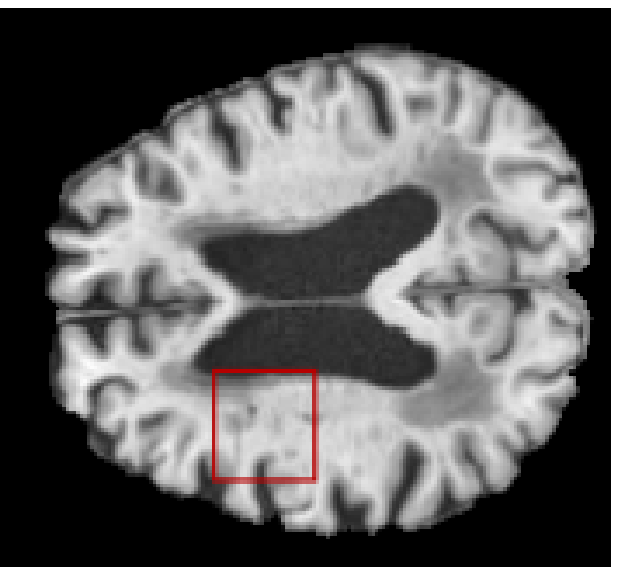}  & \includegraphics[width=9em]{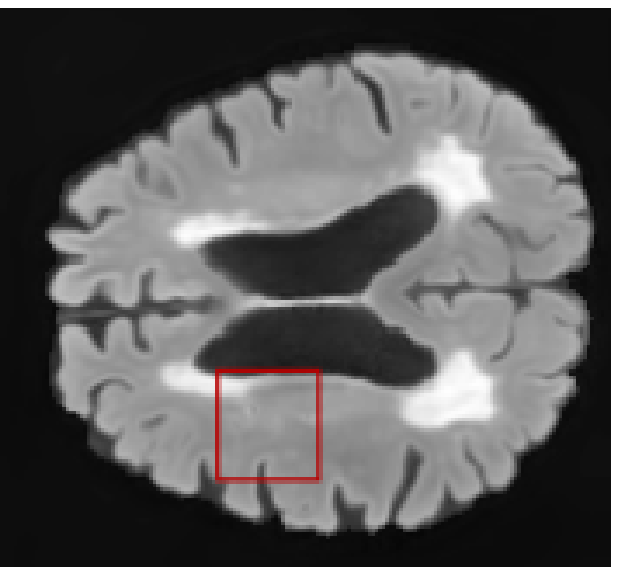} &  \includegraphics[width=9em]{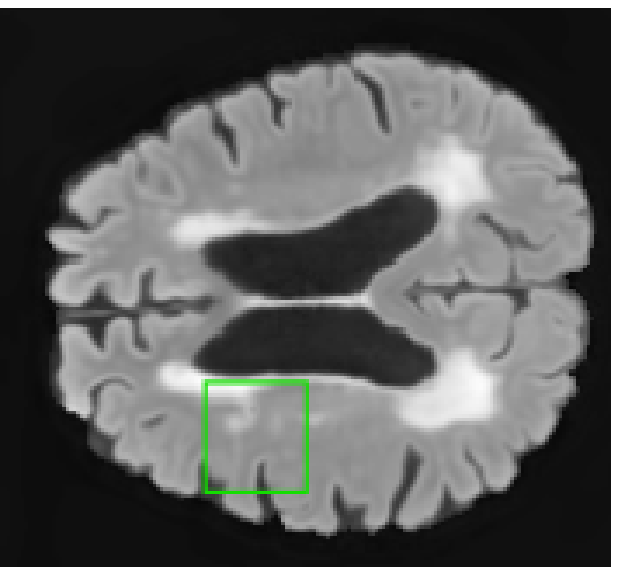}\\
\end{tabular}
\caption{Results of Generation for all the proposed methods, }\label{fig:Reconstructions}
\end{table} 

\section{Discussion and Concluding Remarks}
In this paper, a new CNN-based method to improve WMH segmentation from T1-w images alone is proposed. The method jointly performs imputation and segmentation in such a way that both tasks are mutually benefited.  To this end, FLAIR sequences are used to drive the optimization, which reflects in the results where joint optimization of synthesis and segmentation yield better segmentation from T1-only images. 

From segmentation results in \Cref{segRes}, it is evident that the T1-based segmentation tends to have excessive over-segmentation of images. By using prior information from FLAIR images through a generator, we are able to reduce the number of false positives. 
However, it could be observed that, if imputation comes from an independent synthesis model, images tend to be under segmented (high FNR)  reducing the overall segmentation accuracy. The proposed joint optimization strategy better adapts to capture small lesions, which leads to significantly better overall segmentation performance.

In addition to an improved segmentation performance, we can see in ~\Cref{resGen} that the proposed method also produces better synthetic FLAIR images when compared to networks that trained to only specialize in generation. This may be due to the complementary information available through a joint optimization with the segmentation network. Specially, lesions that are barely visible in T1 images are seen in synthetic images produced by the proposed method. 

One of the disadvantages of our method is using L2 as a loss function can produce blurring effect on the images. Using adversarial training by the use of a discriminative network as a loss function may overcome this issue. However, with an introduction of an additional network and the availability of limited training data, the optimization may be prone overfitting. Therefore the proposed method with L2 loss provides a good compromise between the complexity of the model and segmentation performance. 
\section*{Acknowledgments}
This project has received funding from the European Union’s Horizon 2020 research and innovation
programme under the Marie Skłodowska-Curie grant agreement No 721820. We would like to thank both Microsoft and NVIDIA for providing computational resources on the Azure platform for this project.

\bibliography{Mauro}

\begin{thebibliography}{14}
\providecommand{\natexlab}[1]{#1}
\providecommand{\url}[1]{\texttt{#1}}
\expandafter\ifx\csname urlstyle\endcsname\relax
  \providecommand{\doi}[1]{doi: #1}\else
  \providecommand{\doi}{doi: \begingroup \urlstyle{rm}\Url}\fi

\bibitem[Ben-Cohen et~al.(2018)Ben-Cohen, Klang, Raskin, Soffer, Ben-Haim,
  Konen, Amitai, and Greenspan]{ben2018cross}
Avi Ben-Cohen, Eyal Klang, Stephen~P Raskin, Shelly Soffer, Simona Ben-Haim,
  Eli Konen, Michal~Marianne Amitai, and Hayit Greenspan.
\newblock Cross-modality synthesis from ct to pet using fcn and gan networks
  for improved automated lesion detection.
\newblock \emph{arXiv preprint arXiv:1802.07846}, 2018.

\bibitem[Caligiuri et~al.(2015)Caligiuri, Perrotta, Augimeri, Rocca, Quattrone,
  and Cherubini]{caligiuri2015automatic}
Maria~Eugenia Caligiuri, Paolo Perrotta, Antonio Augimeri, Federico Rocca, Aldo
  Quattrone, and Andrea Cherubini.
\newblock Automatic detection of white matter hyperintensities in healthy aging
  and pathology using magnetic resonance imaging: A review.
\newblock \emph{Neuroinformatics}, 13\penalty0 (3):\penalty0 261--276, 2015.

\bibitem[Dadar et~al.(2017)Dadar, Pascoal, Manitsirikul, Misquitta, Tartaglia,
  Brietner, Rosa-Neto, Carmichael, DeCarli, and Collins]{dadar2017validation}
Mahsa Dadar, Tharick~A Pascoal, Sarinporn Manitsirikul, Karen Misquitta,
  Carmela Tartaglia, John Brietner, Pedro Rosa-Neto, Owen Carmichael, Charles
  DeCarli, and D~Louis Collins.
\newblock Validation of a regression technique for segmentation of white matter
  hyperintensities in alzheimer’s disease.
\newblock \emph{IEEE Transactions on Medical Imaging}, 2017.

\bibitem[Goodfellow et~al.(2014)Goodfellow, Pouget-Abadie, Mirza, Xu,
  Warde-Farley, Ozair, Courville, and Bengio]{goodfellow2014generative}
Ian Goodfellow, Jean Pouget-Abadie, Mehdi Mirza, Bing Xu, David Warde-Farley,
  Sherjil Ozair, Aaron Courville, and Yoshua Bengio.
\newblock Generative adversarial nets.
\newblock In \emph{Advances in neural information processing systems}, pages
  2672--2680, 2014.

\bibitem[Griffanti et~al.(2016)Griffanti, Zamboni, Khan, Li, Bonifacio,
  Sundaresan, Schulz, Kuker, Battaglini, Rothwell, et~al.]{griffanti2016bianca}
Ludovica Griffanti, Giovanna Zamboni, Aamira Khan, Linxin Li, Guendalina
  Bonifacio, Vaanathi Sundaresan, Ursula~G Schulz, Wilhelm Kuker, Marco
  Battaglini, Peter~M Rothwell, et~al.
\newblock Bianca (brain intensity abnormality classification algorithm): A new
  tool for automated segmentation of white matter hyperintensities.
\newblock \emph{Neuroimage}, 141:\penalty0 191--205, 2016.

\bibitem[Havaei et~al.(2016)Havaei, Guizard, Chapados, and
  Bengio]{havaei2016hemis}
Mohammad Havaei, Nicolas Guizard, Nicolas Chapados, and Yoshua Bengio.
\newblock Hemis: Hetero-modal image segmentation.
\newblock In \emph{International Conference on Medical Image Computing and
  Computer-Assisted Intervention}, pages 469--477. Springer, 2016.

\bibitem[Huo et~al.(2017)Huo, Xu, Bao, Assad, Abramson, and
  Landman]{huo2017adversarial}
Yuankai Huo, Zhoubing Xu, Shunxing Bao, Albert Assad, Richard~G Abramson, and
  Bennett~A Landman.
\newblock Adversarial synthesis learning enables segmentation without target
  modality ground truth.
\newblock \emph{arXiv preprint arXiv:1712.07695}, 2017.

\bibitem[Kamnitsas et~al.(2017)Kamnitsas, Ledig, Newcombe, Simpson, Kane,
  Menon, Rueckert, and Glocker]{kamnitsas2017efficient}
Konstantinos Kamnitsas, Christian Ledig, Virginia~FJ Newcombe, Joanna~P
  Simpson, Andrew~D Kane, David~K Menon, Daniel Rueckert, and Ben Glocker.
\newblock Efficient multi-scale 3d cnn with fully connected crf for accurate
  brain lesion segmentation.
\newblock \emph{Medical image analysis}, 36:\penalty0 61--78, 2017.

\bibitem[Li et~al.(2018)Li, Jiang, Wang, Zhang, Wang, Zheng, and
  Menze]{li2018fully}
Hongwei Li, Gongfa Jiang, Ruixuan Wang, Jianguo Zhang, Zhaolei Wang, Wei-Shi
  Zheng, and Bjoern Menze.
\newblock Fully convolutional network ensembles for white matter
  hyperintensities segmentation in mr images.
\newblock \emph{arXiv preprint arXiv:1802.05203}, 2018.

\bibitem[Litjens et~al.(2017)Litjens, Kooi, Bejnordi, Setio, Ciompi,
  Ghafoorian, van~der Laak, van Ginneken, and S{\'a}nchez]{litjens2017survey}
Geert Litjens, Thijs Kooi, Babak~Ehteshami Bejnordi, Arnaud Arindra~Adiyoso
  Setio, Francesco Ciompi, Mohsen Ghafoorian, Jeroen~AWM van~der Laak, Bram van
  Ginneken, and Clara~I S{\'a}nchez.
\newblock A survey on deep learning in medical image analysis.
\newblock \emph{arXiv preprint arXiv:1702.05747}, 2017.

\bibitem[Nie et~al.(2017)Nie, Trullo, Lian, Petitjean, Ruan, Wang, and
  Shen]{nie2017medical}
Dong Nie, Roger Trullo, Jun Lian, Caroline Petitjean, Su~Ruan, Qian Wang, and
  Dinggang Shen.
\newblock Medical image synthesis with context-aware generative adversarial
  networks.
\newblock In \emph{International Conference on Medical Image Computing and
  Computer-Assisted Intervention}, pages 417--425. Springer, 2017.

\bibitem[Tran et~al.(2017)Tran, Pham, Carneiro, Palmer, and
  Reid]{tran2017bayesian}
Toan Tran, Trung Pham, Gustavo Carneiro, Lyle Palmer, and Ian Reid.
\newblock A bayesian data augmentation approach for learning deep models.
\newblock In \emph{Advances in Neural Information Processing Systems}, pages
  2794--2803, 2017.

\bibitem[van Tulder and de~Bruijne(2015)]{van2015does}
Gijs van Tulder and Marleen de~Bruijne.
\newblock Why does synthesized data improve multi-sequence classification?
\newblock In \emph{International Conference on Medical Image Computing and
  Computer-Assisted Intervention}, pages 531--538. Springer, 2015.

\bibitem[Zhang et~al.(2018)Zhang, Yang, and Zheng]{zhang2018translating}
Zizhao Zhang, Lin Yang, and Yefeng Zheng.
\newblock Translating and segmenting multimodal medical volumes with cycle-and
  shape-consistency generative adversarial network.
\newblock \emph{arXiv preprint arXiv:1802.09655}, 2018.

\end{thebibliography}

\end{document}